\documentclass{article}

\usepackage{arxiv}

\usepackage[utf8]{inputenc} % allow utf-8 input
\usepackage[T1]{fontenc}    % use 8-bit T1 fonts
\usepackage{hyperref}       % hyperlinks
\usepackage{url}            % simple URL typesetting
\usepackage{booktabs}       % professional-quality tables
\usepackage{amsfonts}       % blackboard math symbols
\usepackage{nicefrac}       % compact symbols for 1/2, etc.
\usepackage{microtype}      % microtypography
\usepackage{lipsum}		% Can be removed after putting your text content
\usepackage{graphicx}
\usepackage{natbib}
\usepackage{doi}
\usepackage{amsmath}
\usepackage{listings}
\usepackage{color}
\usepackage{subcaption}
\usepackage{float}
\usepackage{listings}
\usepackage{amsmath}
\usepackage{algorithm}
\usepackage{algpseudocode}

\definecolor{dkgreen}{rgb}{0,0.6,0}
\definecolor{gray}{rgb}{0.5,0.5,0.5}
\definecolor{mauve}{rgb}{0.58,0,0.82}

\title{Pearl: Parallel Evolutionary and Reinforcement Learning Library}

\author{ {\hspace{1mm}Rohan Tangri}\thanks{Department of Electrical and Electronic Engineering, Imperial College London, UK, Correspondence to: Rohan Tangri <rohan.tangri16@imperial.ac.uk>} \\
	\And
	{\hspace{1mm}Danilo P. Mandic}$^*$\\
	\And
	{\hspace{1mm}Anthony G. Constantinides}$^*$\\
}

\renewcommand{\shorttitle}{Pearl}

\hypersetup{
pdftitle={A template for the arxiv style},
pdfsubject={q-bio.NC, q-bio.QM},
pdfauthor={David S.~Hippocampus, Elias D.~Striatum},
pdfkeywords={First keyword, Second keyword, More},
}

\begin{document}
\maketitle
\begin{abstract}
	Reinforcement learning is increasingly finding success across domains where the problem can be represented as a Markov decision process. Evolutionary computation algorithms have also proven successful in this domain, exhibiting similar performance to the generally more complex reinforcement learning. Whilst there exist many open-source reinforcement learning and evolutionary computation libraries, no publicly available library combines the two approaches for enhanced comparison, cooperation, or visualization. To this end, we have created Pearl (\url{https://github.com/LondonNode/Pearl}), an open source Python library designed to allow researchers to rapidly and conveniently perform optimized reinforcement learning, evolutionary computation and combinations of the two. The key features within Pearl include: modular and expandable components, opinionated module settings, Tensorboard integration, custom callbacks and comprehensive visualizations.
\end{abstract}

% keywords can be removed

\section{Introduction}
Reinforcement learning (RL) is a growing area of research with impressive results in robotics (\cite{DBLP:journals/corr/abs-1808-00177}), communications (\cite{8761183}) and finance (\cite{balch2019evaluate}). With the necessity to process increasing volumes of data, the algorithm complexity has also been growing. In addition, different classes of agents tend to have differing fundamentals; for example, off-policy SAC (\cite{haarnoja2019soft}) vs on-policy PPG (\cite{pmlr-v139-cobbe21a}). This, in turn, tends to increase complexity in software libraries attempting to unify these ideas in a single structure (\cite{pineau2020improving}). Furthermore, evolutionary computation (EC) is competitive with RL whilst often being algorithmically simpler, although this may come at the cost of data inefficiency (\cite{salimans2017evolution}). Methodologies combining both approaches in hybrid algorithms are also emerging, and have been shown to outperform both individual techniques separately (\cite{majid2021deep}, \cite{DBLP:journals/corr/abs-1810-01222}). This has also highlighted a void in the literature and available libraries, especially given the need for highly modular software allowing researchers to rapidly prototype and test new algorithms in both the EC and RL frameworks for comparison and cooperation.

To this end, we introduce Pearl (\textbf{P}arallel \textbf{E}volutionary \textbf{a}nd \textbf{R}einfocement \textbf{L}earning Library), a Python library designed for rapid prototyping and testing of adaptive decision making algorithms, including tools for both RL and EC. Supported use cases are as follows: 1) Single agent RL 2) Multi-agent RL 3) EC in a trajectory environment with observations, actions and rewards 4) EC for optimizing black box static functions (useful for direct testing of EC algorithm properties in differently shaped functions) 5) Hybrid algorithms combining RL and EC to optimize policies. The deep learning components are built in PyTorch (\cite{NEURIPS2019_9015}) while further details of other dependencies can be found in the \texttt{pyproject.toml} documentation. The Pearl package provides modular and extensible components which can be plugged into compatible agents for ease of experimentation and resilience to code changes. Opinionated settings dataclasses are also used to group together module parameters and give user suggestions. In addition, custom callbacks are implemented to allow users to inject unique logic into their agents during training (\cite{stable-baselines3}). Finally, Pearl offers Tensorboard integration for real time training analysis, while executable scripts introduce a basic command line interface (CLI) for the visualization of complex results and basic demonstrations of implemented agents.

\section{Library Design}
\label{sec:library_design}

\begin{figure}[h]
    \centering
    \includegraphics[width=15cm]{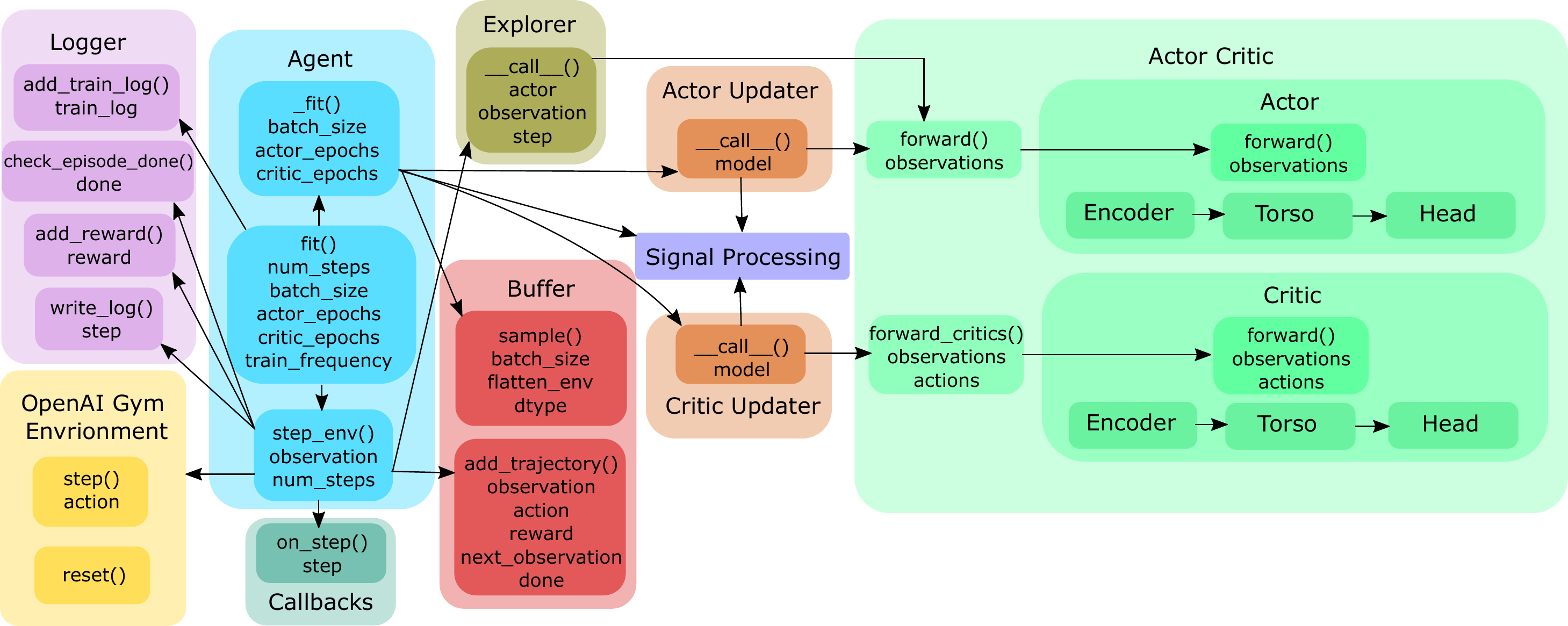}
    \caption{Module flow for training an agent}
    \label{fig:1}
\end{figure}

Pearl minimizes the library learning curve by allowing EC, RL and hybrid agents to be derived from a single base class. The high level training flow for an agent derived from the base class is illustrated in Figure \ref{fig:1}. The structure of Pearl includes the following modules:
\begin{itemize}
    \item {The \textbf{models} represent neural network structures. An actor-critic object is provided that accepts a separate or shared actor and critic network as a template. This template can then be used to generate a population of actor and critic networks. Methods are provided for an easy forward pass through either population. An individual network can also be represented as a vector of network parameters, so the populations of actors and critics can each be represented by a matrix, and a matrix can be used to update each population. Finally, there is the notion of a global network, which acts as the average of the population networks to be used post-training. As shown in Figure \ref{fig:1}, each actor and critic is designed as a sequence of an encoder, a torso and a head. The encoder processes the input; for example, by concatenating the state observation and action for the continuous Q function network. The torso embodies the deep layers; in our case, a multilayer perceptron. Finally, the head dictates the output; for example, a categorical distribution for an actor. Each actor and critic can also be specified to have an associated target network, which will scale with the population in the actor-critic object. Furthermore, a dummy model is provided that doesn't utilize any deep structure but can still be used in the actor-critic object to allow for EC in static black box optimization tasks.}
    
    \item {The \textbf{agents} combine the modules within Pearl, and act as the interface between the user and the algorithm. The benefit of this single interface is a high level of abstraction. Generally, this comes at the cost of a potentially confusing long list of inputs (\cite{pardo2020tonic}), but within Pearl this is avoided through the use of settings dataclasses which group together the various module parameters and give user suggestions on what inputs are needed and their types. Such benefits are not found in dictionary objects which are often used for this task. Implementations of EC, RL and hybrid agents are provided as examples.}
    
    \item {The \textbf{buffers} handle the storing and sampling of agent trajectories in the environment. To promote flexibility, all trajectories in the buffer can be collected, or alternatively, they can be randomly sampled, or only the last $n$ trajectories can be selected. For off-policy buffers, a single data structure is used to store both trajectory observations and next observations, thus halving memory requirements. This stems from the sequential nature of observations with the information 'blended' only at episode end.}

    \item {The \textbf{updaters} handle iterative updates for neural networks and any other adaptive or iterative models. There are three classes of updater: actor updaters, critic updaters and evolutionary updaters.}
    
    \item {The \textbf{signal processing} module designates the flow of information in both RL and EC, with methods such as the generalized advantage estimate (\cite{schulman2018highdimensional}) or sample approximations for KL divergence (\cite{schulman_2021}). This further enhances modularity, since these are implemented as procedural methods which can be straightforwardly used for experimenting with different updater algorithms.}
    
    \item {The \textbf{explorers} are responsible for processing actions given by a deterministic actor network, and thus improve exploration. This mainly involves adding noise to algorithms such as DDPG. Another useful feature is the ability to randomly explore the environment, independent of the actor network, for the first $n$ steps of training.}
    
    \item {The \textbf{logger} handles useful training statistics which are written to Tensorboard and a log file. By default, the Tensorboard logs are split into three sections: the reward section monitors the rewards the agent receives, the loss section monitors the actor and critic loss values, and the metrics section monitors useful metrics such as the policy entropy or KL divergence. This is particularly useful for monitoring model convergence.}
    
    \item {The \textbf{callbacks} are designed to allow the user to inject unique logic into the training loop at every time step. This allows for useful features, such as saving the current model and early stopping.}
\end{itemize}

\section{Visualization and Scripts}
\begin{figure}[h]
     \centering
     \begin{subfigure}{.59\textwidth}
         \centering
         \includegraphics[width=\textwidth]{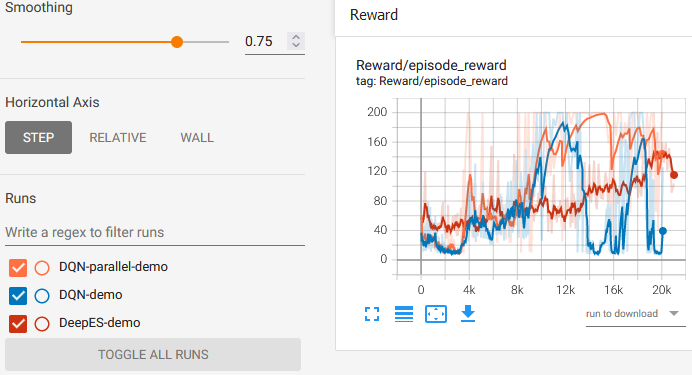}
         \caption{Tensorboard live training monitor}
         \label{fig:tensorboard}
     \end{subfigure}
     \hfill
     \begin{subfigure}{.4 \textwidth}
         \centering
         \includegraphics[width=\textwidth]{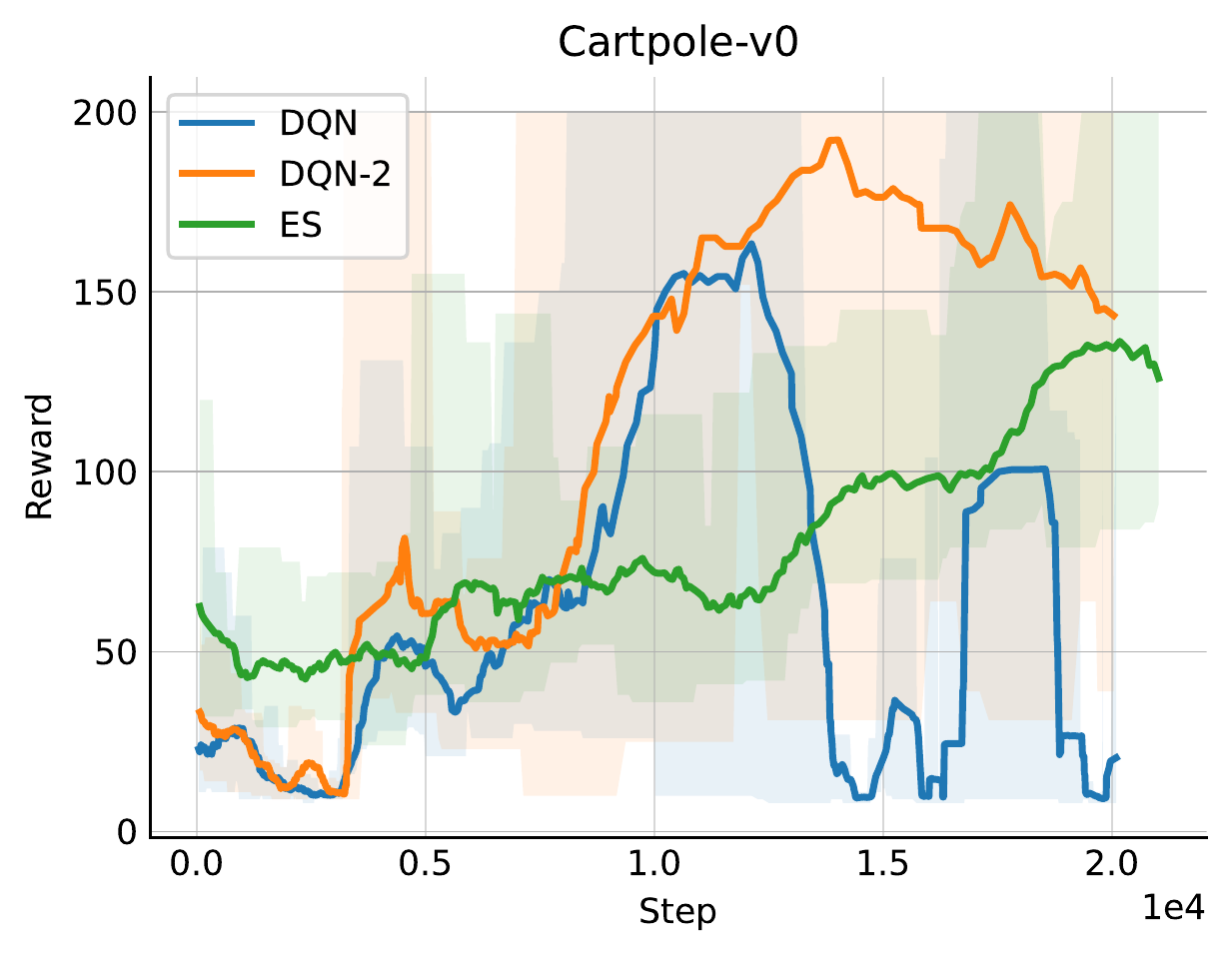}
         \caption{Post training plot via CLI}
         \label{fig:plot}
     \end{subfigure}
     \caption{Visualization capabilities within Pearl}
\end{figure}

Pearl provides manifold ways to visualize results, such as live analysis via Tensorboard, as shown in Figure \ref{fig:tensorboard}. This is useful for monitoring progress in long training runs. To accommodate more complicated plots, for example, comparing different runs with plot titles, axes titles and legends, a plotting script can be executed via a CLI. An exemplar plot is shown in Figure \ref{fig:plot}, which is created by the command: \texttt{python -m pearll.plot -p runs/DQN-demo runs/DQN-parallel-demo runs/DeepES-demo ---metric reward ---titles Cartpole-v0 ---num-cols 1 ---legend DQN DQN-2 ES ---window 20}.

Furthermore, a demo script is also provided that allows users to quickly demonstrate agents in simple environments without the need to open any Python editors. For example, the command \texttt{python -m pearll.demo ---agent her} will run a DQN agent with the HER buffer in a discrete bit flipping environment (\cite{DBLP:journals/corr/AndrychowiczWRS17}). These are useful to quickly verify that agents are working as expected and act as integration tests.

\section{AdamES}

To demonstrate the utility of Pearl for rapidly prototyping, testing and visualizing new RL and EC algorithms, we implement AdamES, a novel EC algorithm which aims to combine OpenAI's Natural Evolutionary Strategy (ES) (\cite{salimans2017evolution}) with the Adam optimizer equation (\cite{Kingma2015AdamAM}).

\subsection{Algorithm}

Suppose there exists a continuous, differentiable function, $F(\theta)$. The ES equation mirrors the standard gradient descent equation by making a sample gradient approximation (\cite{salimans2017evolution}). Meanwhile, the optimization domain has extended stochastic gradient descent through the use of momentum (\cite{QIAN1999145}) and dampening (\cite{dauphin2015equilibrated}) for improved stability and convergence speed; however, these concepts are not applied in the ES algorithm. In AdamES, we replace the gradient terms of the Adam optimizer with the ES approximation. In this way, we add the concepts of momentum and dampening to the update equation to increase convergence speed and steady state stability for a minor algorithmic change. The code for the AdamES algorithm within the Pearl framework is in Appendix \ref{app:AdamES}.

\begin{algorithm}[H]
\caption{\textit{AdamES}, all operations on vectors are element-wise.}\label{alg:AdamES}
\begin{algorithmic}
\State \textbf{Input:} initial policy parameters $\theta_0$, momentum exponential weight $\beta_1 \in [0, 1]$, dampening exponential weight $\beta_2 \in [0, 1]$, sampling noise standard deviation $\sigma$, population size $n$, learning rate $\alpha$
\State momentum term $m_0 \gets 0$
\State dampening term $v_0 \gets 0$
\For{$t = 0, 1, 2, \dotsc, T-1$}
\State Sample $\epsilon_1, \dotsc, \epsilon_n \sim \mathcal{N}(0, I)$
\State $R_i = F(\theta_t + \sigma \epsilon_i)$ for $i=1, \dotsc, n$ \Comment{Compute population returns}
\State $\nabla F(\theta_t) \approx \frac{1}{n \sigma} \sum_{i=1}^n R_i\epsilon_i$ \Comment{Compute gradient approximation}
\State $m_{t+1} \gets (1-\beta_1)\nabla F(\theta_t)  + \beta_1m_{t}$ \Comment{Update momentum term}
\State $v_{t+1} \gets (1-\beta_2) [\nabla F(\theta_t) \odot \nabla F(\theta_t)] + \beta_2v_{t}$ \Comment{Update dampening term}
\State $\hat{m}_{t+1} \gets m_{t+1}/(1-(\beta_1)^t)$
\State $\hat{v}_{t+1} \gets v_{t+1}/(1-(\beta_2)^t)$
\State $\theta_{t+1} \gets \theta_t + [\alpha \hat{m}_{t+1}/(\sqrt{\hat{v}_{t+1}} + \gamma)]$ \Comment{$\gamma=$ fuzz term to avoid divide by 0 error}
\EndFor
\State \textbf{Output:} optimized policy $\theta_T$
\end{algorithmic}
\end{algorithm}

\subsection{Experiments}

\begin{figure}[h]
    \centering
    \includegraphics[width=\textwidth]{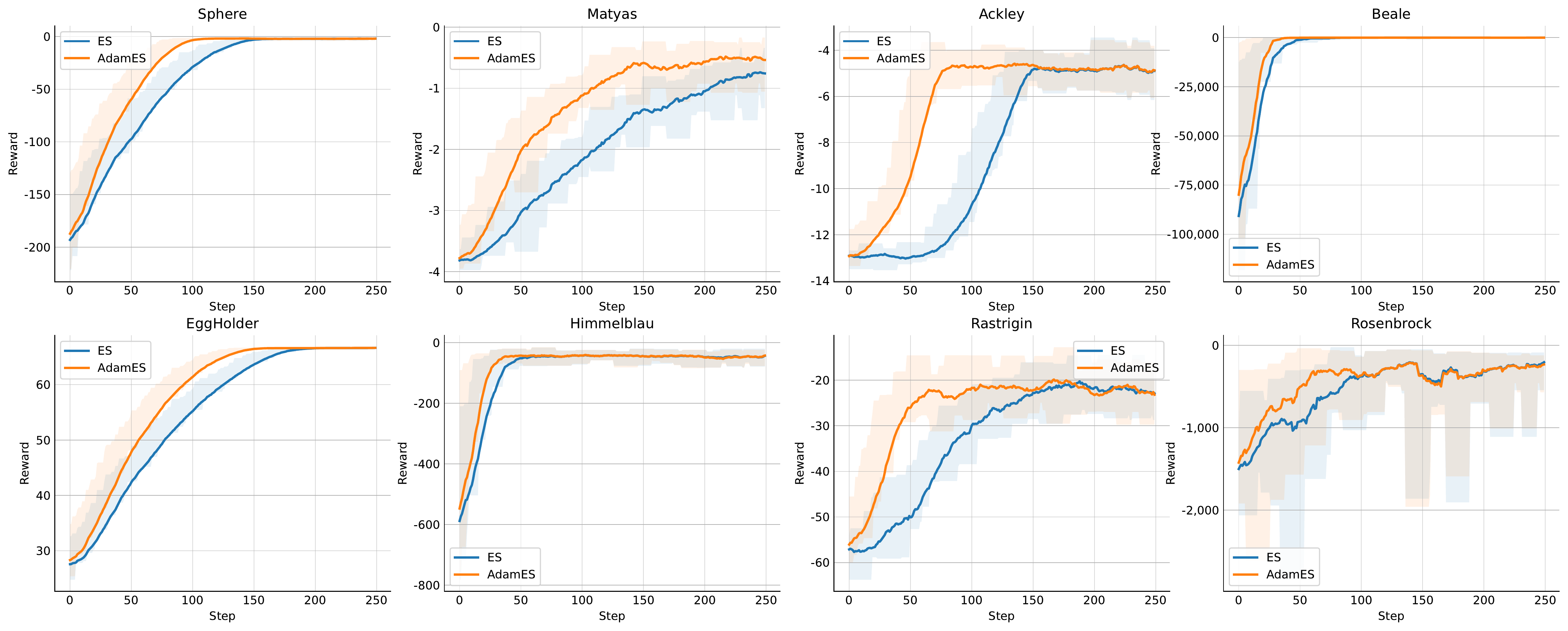}
    \caption{Learning curves}
    \label{fig:4}
\end{figure}

\begin{figure}[h]
    \centering
    \includegraphics[width=\textwidth]{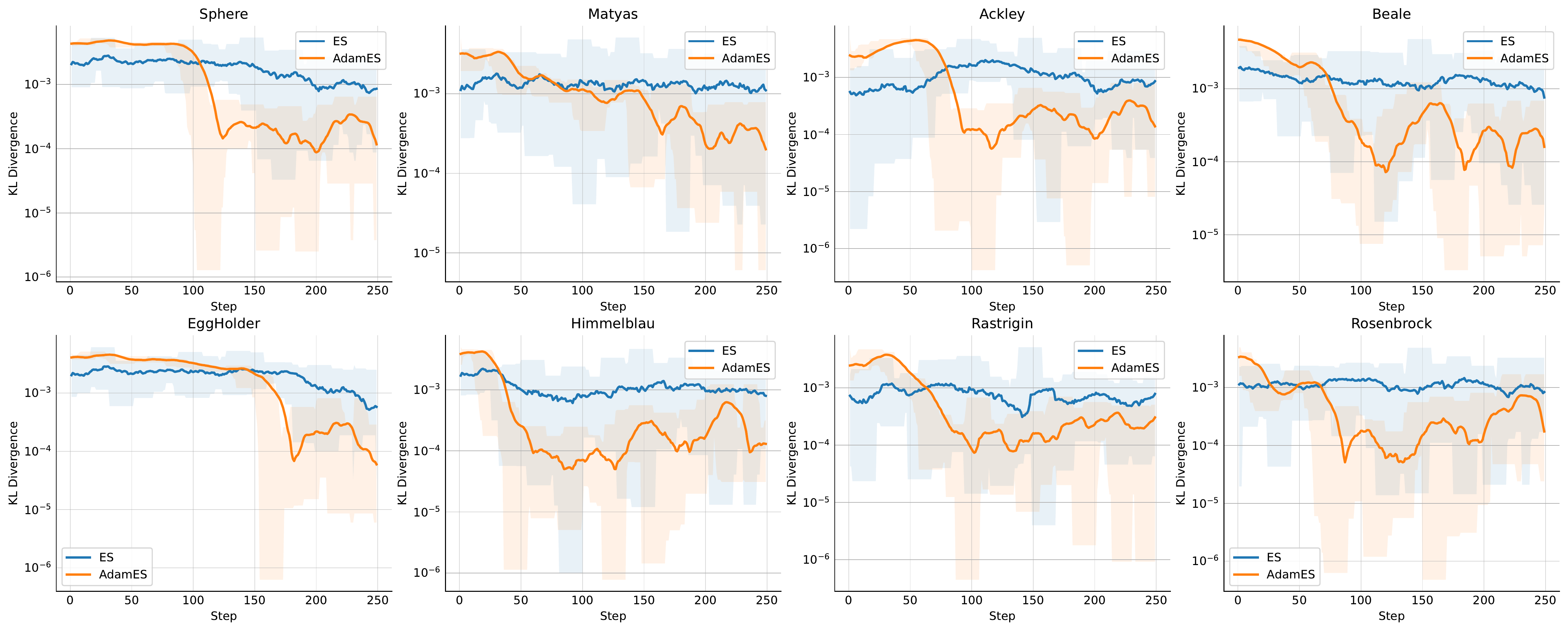}
    \caption{Population divergence with log scale}
    \label{fig:5}
\end{figure}

The learning curves of AdamES and ES on a set of multidimensional continuous optimization functions are shown in Figure \ref{fig:4} and the KL divergence between population iterations are shown in Figure \ref{fig:5}. Pearl allows for easy and rapid generation of these plots via the CLI. Different metrics logged by default via Tensorboard can be plotted using the \texttt{---metric} flag, which can be set to plot rewards, divergences, entropies, actor losses or critic losses. Furthermore, a log scale can be used for the y-axis by including the flag \texttt{---log-y}. Finally, the plot name and file types can be set by the commands \texttt{---save-path} and \texttt{---save-types}, which allows the same plot to be saved in many different formats, by default, a PDF.

Hyperparameters used in the experiments plotted in Figure \ref{fig:4} and Figure \ref{fig:5} are listed in Appendix \ref{app:params}. AdamES achieves faster convergence speed as shown directly by the learning curves and the larger KL divergence between population iterations while the algorithm has not reached a maximum. At the same time, while the steady state KL divergence of AdamES has a higher variance than the steady state KL divergence of ES, it also has a smaller expected value. This indicates that AdamES has a greater stability around the solution found compared to ES.

\newpage

\section{Conclusions and Future Work}
This paper has introduced Pearl, a library especially designed to allow researchers to rapidly prototype and test new ideas in order to optimize decision making algorithms in reinforcement learning type environments. Through easy access to interpretable tools, Pearl facilitates experimentation, further research and innovation, particularly at the overlap between RL and EC. 

Future work will involve new agents and features such as the Intrinsic Curiosity Module (\cite{pathakICMl17curiosity}) as well as the expansion of the models module to other types of neural networks, such as transformers (\cite{vaswani2017attention}).

\newpage
\bibliographystyle{unsrtnat}
\bibliography{template}  %%% Uncomment this line and comment out the ``thebibliography'' section below to use the external .bib file (using bibtex) .

%\begin{thebibliography}{1}

%	\bibitem{kour2014real}
%	George Kour and Raid Saabne.
%	\newblock Real-time segmentation of on-line handwritten arabic script.
%	\newblock In {\em Frontiers in Handwriting Recognition (ICFHR), 2014 14th
%			International Conference on}, pages 417--422. IEEE, 2014.
%
%	\bibitem{kour2014fast}
%	George Kour and Raid Saabne.
%	\newblock Fast classification of handwritten on-line arabic characters.
%	\newblock In {\em Soft Computing and Pattern Recognition (SoCPaR), 2014 6th
%			International Conference of}, pages 312--318. IEEE, 2014.
%
%	\bibitem{hadash2018estimate}
%	Guy Hadash, Einat Kermany, Boaz Carmeli, Ofer Lavi, George Kour, and Alon
%	Jacovi.
%	\newblock Estimate and replace: A novel approach to integrating deep neural
%	networks with existing applications.
%	\newblock {\em arXiv preprint arXiv:1804.09028}, 2018.
%
%\end{thebibliography}

\newpage

\appendix
\section{AdamES Code}
\label{app:AdamES}

\begin{lstlisting}
class AdamES(BaseAgent):
    def __init__(
        self,
        env: VectorEnv,
        model: ActorCritic,
        updater_class: Type[BaseEvolutionUpdater] = NoisyGradientAscent,
        learning_rate: float = 1,
        momentum_weight: float = 0.9,
        damping_weight: float = 0.999,
        buffer_class: Type[BaseBuffer] = RolloutBuffer,
        buffer_settings: BufferSettings = BufferSettings(),
        action_explorer_class: Type[BaseExplorer] = BaseExplorer,
        explorer_settings: ExplorerSettings = ExplorerSettings(start_steps=0),
        callbacks: Optional[List[Type[BaseCallback]]] = None,
        callback_settings: Optional[List[CallbackSettings]] = None,
        logger_settings: LoggerSettings = LoggerSettings(),
        device: Union[T.device, str] = "auto",
        render: bool = False,
        seed: Optional[int] = None,
    ) -> None:
        super().__init__(
            env=env,
            model=model,
            action_explorer_class=action_explorer_class,
            explorer_settings=explorer_settings,
            buffer_class=buffer_class,
            buffer_settings=buffer_settings,
            logger_settings=logger_settings,
            callbacks=callbacks,
            callback_settings=callback_settings,
            device=device,
            render=render,
            seed=seed,
        )

        self.learning_rate = learning_rate
        self.momentum_weight = momentum_weight
        self.damping_weight = damping_weight
        self.updater = updater_class(model=self.model)
        self.m = 0
        self.v = 0
        self.adam_step = 1

    def _adam(self, grad: np.floating) -> np.floating:
        """Adam optimizer update"""
        self.m = (1 - self.momentum_weight) * grad + self.momentum_weight * self.m
        self.v = (1 - self.damping_weight) * (
            grad * grad
        ) + self.damping_weight * self.v
        m_adj = self.m / (1 - (self.momentum_weight ** self.adam_step))
        v_adj = self.v / (1 - (self.damping_weight ** self.adam_step))
        self.adam_step += 1
        return m_adj / (np.sqrt(v_adj) + 1e-8)

    def _fit(
        self, batch_size: int, actor_epochs: int = 1, critic_epochs: int = 1
    ) -> Log:
        divergences = np.zeros(actor_epochs)
        entropies = np.zeros(actor_epochs)

        trajectories = self.buffer.all(flatten_env=False)
        rewards = trajectories.rewards.squeeze()
        rewards = filter_rewards(rewards, trajectories.dones.squeeze())
        if rewards.ndim > 1:
            rewards = rewards.sum(axis=-1)
        scaled_rewards = scale(rewards)
        grad_approx = np.dot(self.updater.normal_dist.T, scaled_rewards) / (
            np.mean(self.updater.std) * self.env.num_envs
        )
        optimization_direction = self._adam(grad_approx)
        for i in range(actor_epochs):
            log = self.updater(
                learning_rate=self.learning_rate,
                optimization_direction=optimization_direction,
            )
            divergences[i] = log.divergence
            entropies[i] = log.entropy
        self.buffer.reset()

        return Log(divergence=divergences.sum(), entropy=entropies.mean())
\end{lstlisting}

\section{Hyperparameters}
\label{app:params}

\begin{table}[H]
\centering
\begin{tabular}{|ll|}
\hline
n       & 10    \\ \hline
$\sigma$      & 1     \\ \hline
$\alpha$     & 0.1   \\ \hline
$\beta_1$ & 0.9   \\ \hline
$\beta_2$ & 0.999 \\ \hline
seed & 0 \\ \hline
\end{tabular}
\end{table}

\section{Learning Rate Performance}
\label{app:lr}

\begin{figure}[h]
    \centering
    \includegraphics[width=16cm]{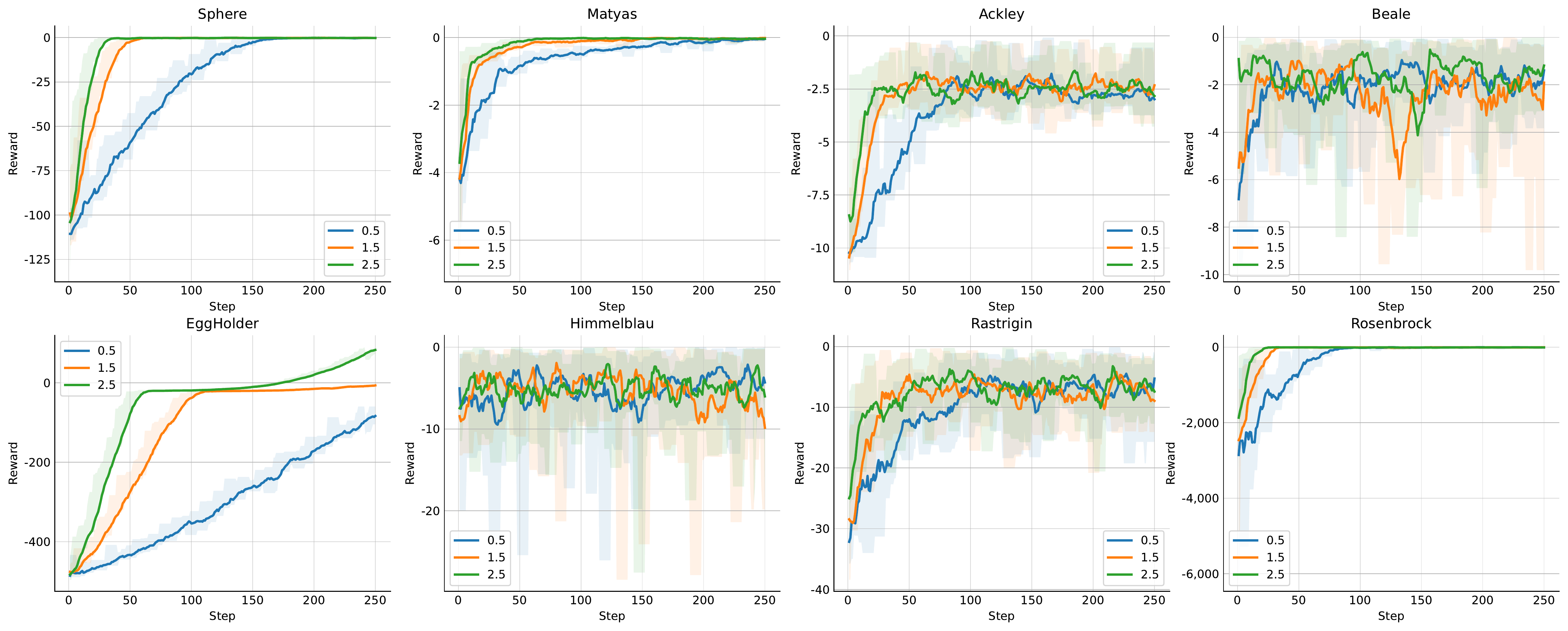}
    \caption{Learning curves with different learning rates}
    \label{fig:6}
\end{figure}

\begin{figure}[h]
    \centering
    \includegraphics[width=16cm]{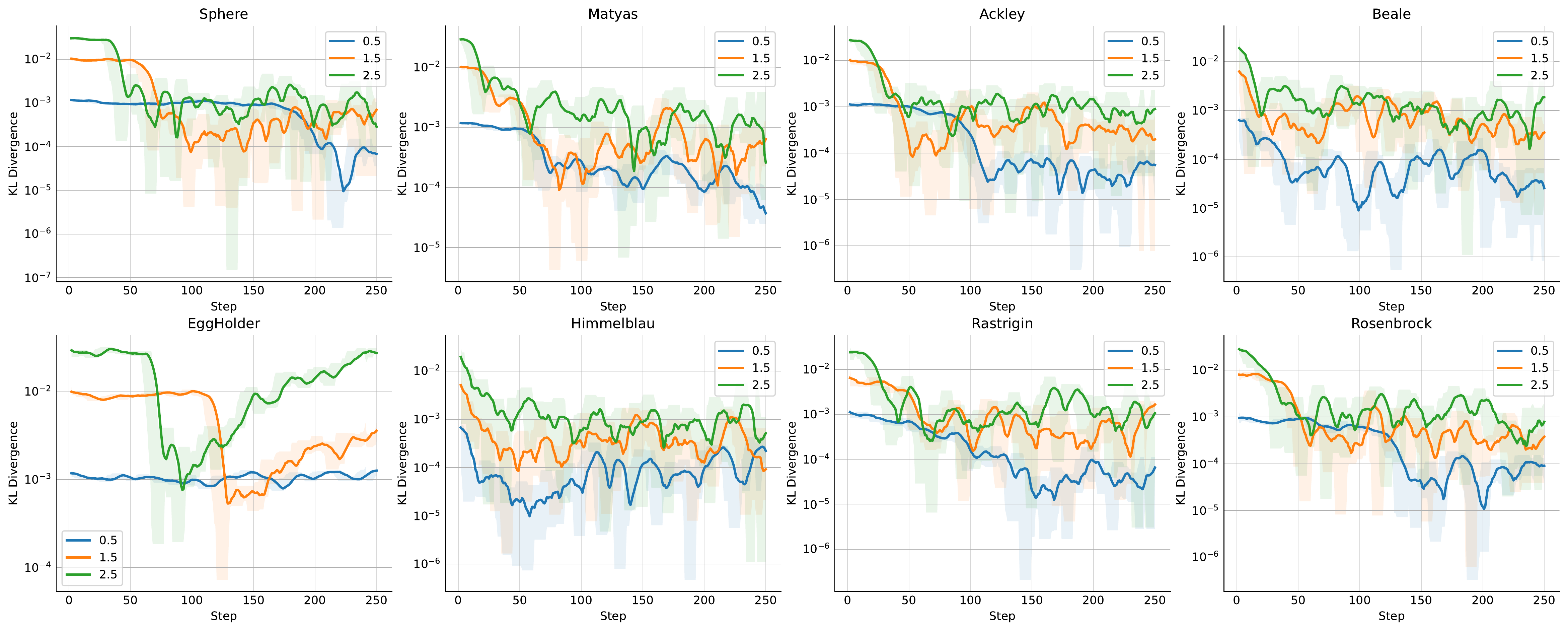}
    \caption{Population divergence with log scale and different learning rates}
    \label{fig:7}
\end{figure}

Figure \ref{fig:6} and Figure \ref{fig:7} show how the learning curves and population KL divergences change as learning rate, $\alpha$, is adjusted. The key thing to note is the nonlinear relationship in both convergence speed and steady state divergence.

\section{Sampling Noise Performance}
\label{app:std}

\begin{figure}[h]
    \centering
    \includegraphics[width=16cm]{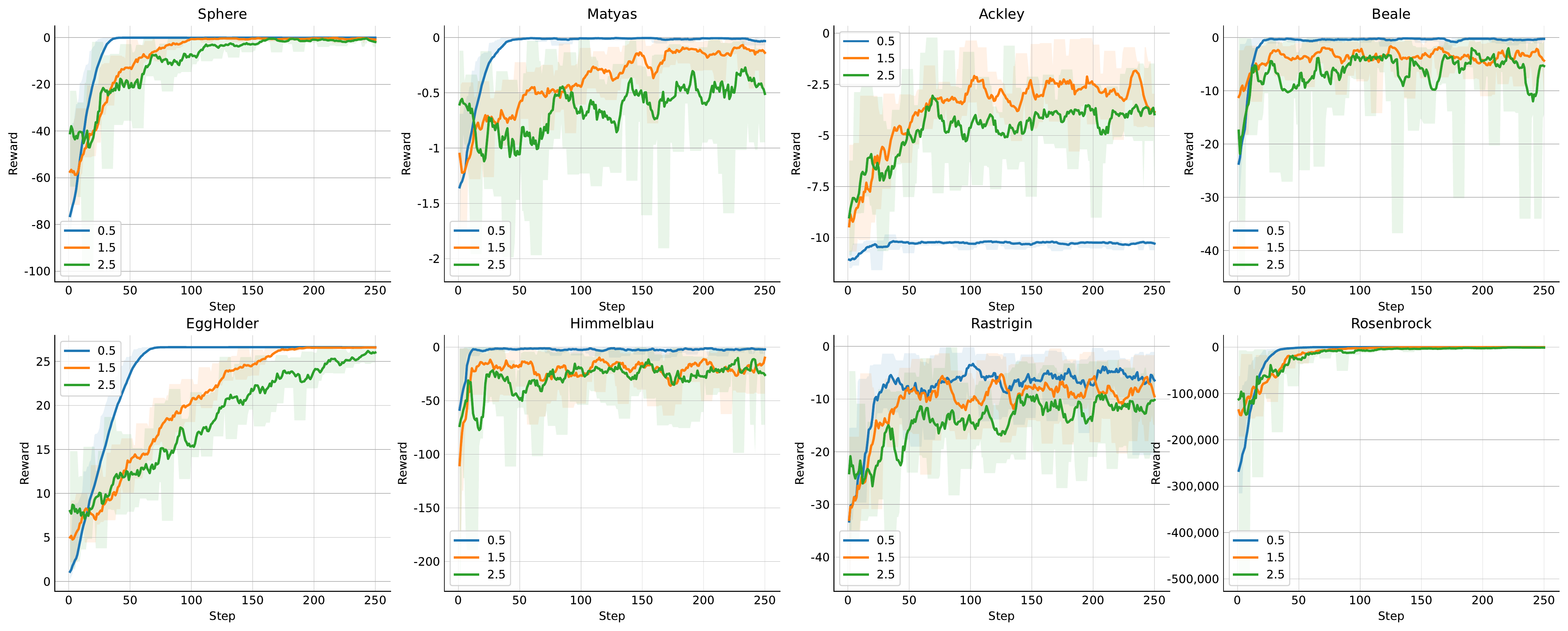}
    \caption{Learning curves with different population sampling noise}
    \label{fig:8}
\end{figure}

\begin{figure}[h]
    \centering
    \includegraphics[width=16cm]{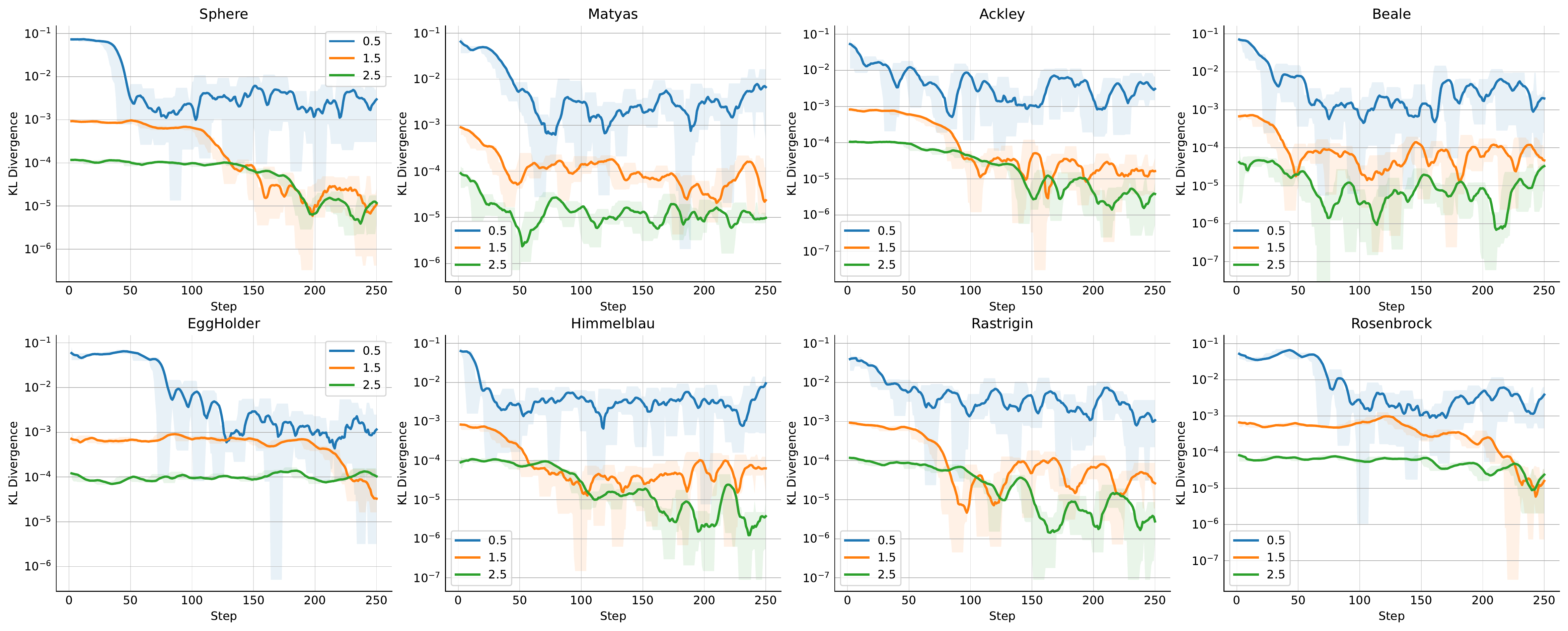}
    \caption{Population divergence with log scale and different population sampling noise}
    \label{fig:9}
\end{figure}

Figure \ref{fig:8} and Figure \ref{fig:9} show how the learning curves and population KL divergences change as population sampling noise standard deviation, $\sigma$, is adjusted. Because the population size is kept constant, the larger sampling noise standard deviation leads to reduced information density and more noisy  gradient ascent approximations. This results in slower convergence and worse steady state divergence. However, it is often useful to increase this parameter to help avoid local traps. When this is necessary, the population size should also be increased to maintain information density.

\section{Population Size Performance}
\label{app:pop}

\begin{figure}[h]
    \centering
    \includegraphics[width=16cm]{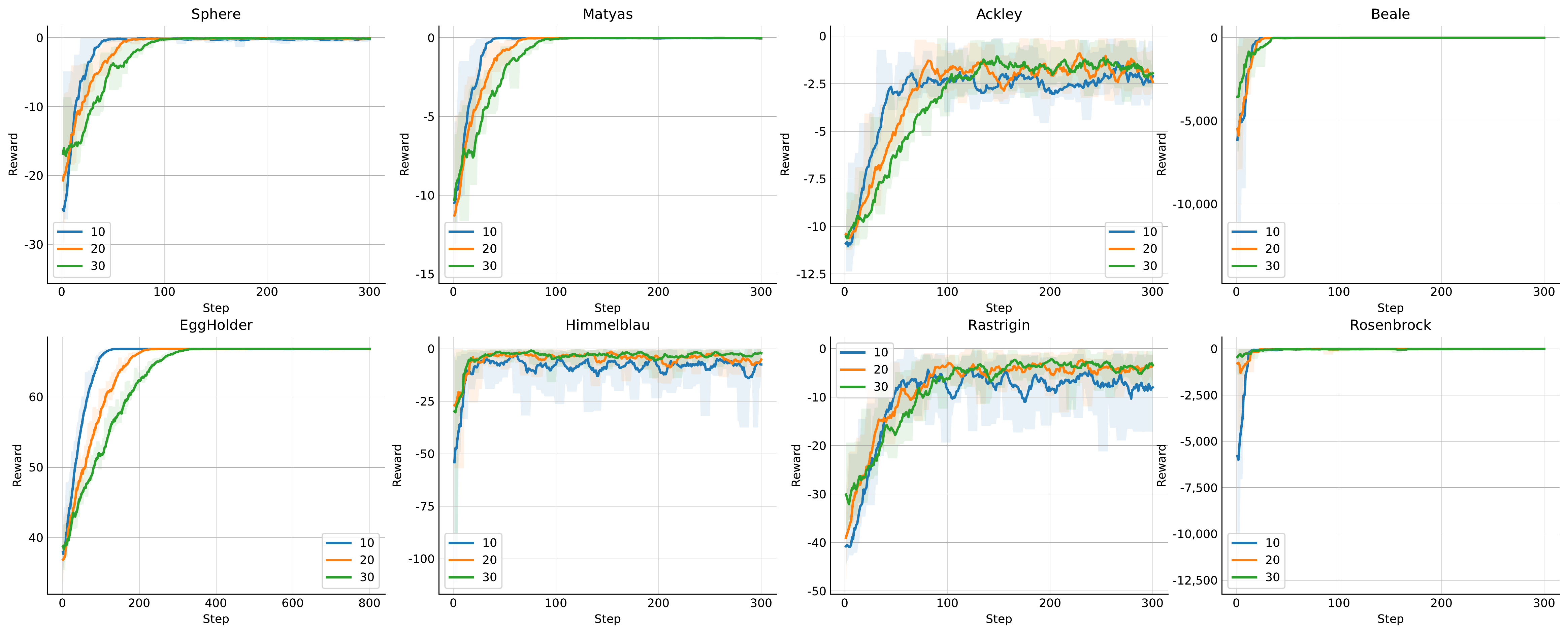}
    \caption{Learning curves with different population size}
    \label{fig:10}
\end{figure}

\begin{figure}[h]
    \centering
    \includegraphics[width=16cm]{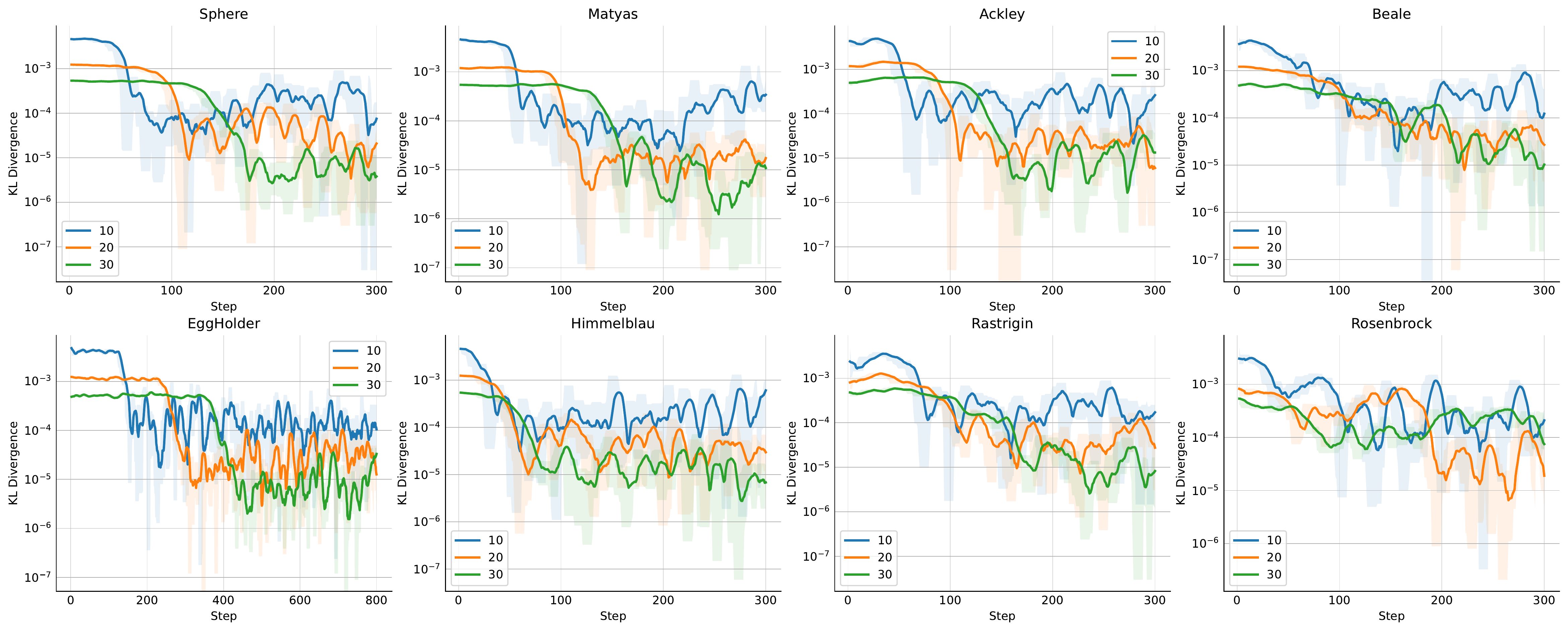}
    \caption{Population divergence with log scale and different population size}
    \label{fig:11}
\end{figure}

Figure \ref{fig:10} and Figure \ref{fig:11} show how the learning curves and population KL divergences change as population size, $n$, is adjusted. With a constant sampling noise, increasing population size increases the information density of sampling, leading to better steady state dynamics. The effect on convergence speed is dependent on the function being optimized due to the ratio of gradient approximations used in the momentum and dampening terms.

\end{document}